\pdfoutput=1

\documentclass[11pt]{article}
\usepackage[final]{acl}

\usepackage{times}
\usepackage{latexsym}

\usepackage{listings}  %

\usepackage[T1]{fontenc}

\usepackage[utf8]{inputenc}

\usepackage{microtype}

\usepackage{inconsolata}

\usepackage{graphicx}
\usepackage{algorithm}
\usepackage{amsmath}
\usepackage{amssymb}
\usepackage{algpseudocode}

\usepackage{booktabs}
\usepackage{colortbl}
\usepackage{pifont}
\usepackage{multirow}
\usepackage{makecell}
\usepackage{mathrsfs}
\usepackage{paralist}
\usepackage{enumitem}
\usepackage{cleveref}
\usepackage{arydshln}
\usepackage{tcolorbox}
\usepackage{multicol}
\tcbuselibrary{skins}
\tcbuselibrary{breakable}  

\definecolor{lightpurple}{HTML}{E8E8FF}

\title{
    SCOPE: Compress Mathematical Reasoning Steps for Efficient Automated Process Annotation
}

\author{
  Huimin Xu$^1$, \quad Xin Mao$^1$, \quad Feng-Lin Li$^2$, \quad Xiaobao Wu$^1$\thanks{Corresponding Authors.} \\
  \textbf{Wang Chen}$^2$, \quad \textbf{Wei Zhang}$^{3}$, \quad \textbf{Anh Tuan Luu}$^{1}$\footnotemark[1] \\
  $^1$Nanyang Technological University, Singapore \\
  $^2$Shopee Pte. Ltd, Singapore, \quad  $^3$SEA Group, Singapore\\\
  \texttt{\{huimin.xu, xin.mao, xiaobao.wu, anhtuan.luu\}@ntu.edu.sg} \\ 
  \texttt{\{fenglin.li, chen.wang\}@shopee.com}, \texttt{terry.zhang@sea.com}
}

\begin{document}
\maketitle
\begin{abstract}
Process Reward Models (PRMs) have demonstrated promising results in mathematical reasoning, but existing process annotation approaches, whether through human annotations or Monte Carlo simulations, remain computationally expensive. In this paper, we introduce Step COmpression for Process Estimation (SCOPE), a novel compression-based approach that significantly reduces annotation costs. We first translate natural language reasoning steps into code and normalize them through Abstract Syntax Tree, then merge equivalent steps to construct a prefix tree. Unlike simulation-based methods that waste numerous samples on estimation, SCOPE leverages a compression-based prefix tree where each root-to-leaf path serves as a training sample, reducing the complexity from $O(NMK)$ to $O(N)$. We construct a large-scale dataset containing 196K samples with only 5\% of the computational resources required by previous methods. Empirical results demonstrate that PRMs trained on our dataset consistently outperform existing automated annotation approaches on both Best-of-N strategy and ProcessBench~\footnote{Our code, data, and models are available at \url{https://github.com/Anna7355/SCOPE}.}.
\end{abstract}

\section{Introduction}
\label{sec:intro}

As Large Language Models (LLMs) advance in complex reasoning tasks \cite{jaech2024openai, liu2024deepseek, yang2024qwen2, dubey2024llama}, designing effective reward models has become increasingly crucial. 
Process Reward Models (PRMs) \cite{uesato2022solving,lightman2023let} evaluate the reasoning process step-by-step, providing more fine-grained supervision than Outcome Reward Models (ORMs) that only assess final outputs \cite{cobbe2021training, yu2023outcome,wu2025sailing}.
Recent studies consistently demonstrate PRMs' superior performance across complex reasoning tasks \cite{wang2024math,snell2024scaling}.
But PRMs face a significant challenge: they require intensive human effort for data annotation, as every reasoning step needs a label \cite{lightman2023let}.

\begin{figure}[t!]
\centering
\includegraphics[width=0.85\linewidth]{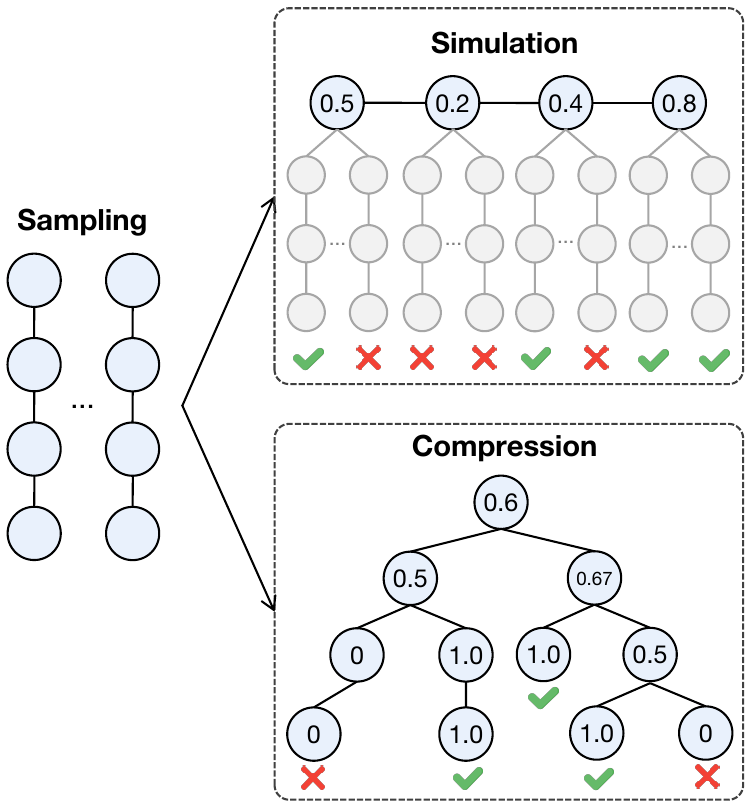}
\caption{Comparison of PRMs training data construction. Simulation-based methods require numerous completions solely for Q-value estimation, with these completions (gray nodes) being discarded without contributing to training. Our compression-based approach eliminates such data waste by merging equivalent steps from all sampled solutions into a prefix tree, where every root-to-leaf path becomes a valuable training instance.}
\label{fig:fig1}
\end{figure}

To alleviate this limitation, recent studies explore simulation-based methods.
For instance, Math-shepherd \cite{wang2024math} uses Monte Carlo estimation to automate the data annotation.
As shown in \Cref{fig:fig1}, it samples $N$ solutions for a math problem, and then for each step in the solution, it simulates $M$ potential completions and calculates the Q-value based on the proportion of completions that lead to the correct answer.
While Math-shepherd eliminates human annotation requirements, it incurs high computational complexity of $O(NMK)$, where $K$ denotes the average step count of solutions. 
Recent studies have shown that Math-shepherd requires $38.8\times$ more FLOPs than ORM training \cite{yuan2024free}. 
Although OmegaPRM \cite{luo2024improve} reduces complexity to $O(NM\log K)$ through a divide-and-conquer strategy, the efficiency gains remain limited due to typically short lengths ($K < 10$). 
We argue that these simulation-based approaches are inherently inefficient as they generate numerous completions solely for Q-value estimation (gray area in \Cref{fig:fig1}), resulting in wasted data.

In this paper, we introduce \textbf{S}tep \textbf{CO}mpression for \textbf{P}rocess \textbf{E}stimation (\textbf{SCOPE}), a novel automatic PRM label annotation strategy that achieves $O(N)$ complexity while maintaining annotation quality.
Unlike prior simulation-based methods \cite{wang2024math, luo2024improve}, SCOPE introduces a novel compression-based approach: first samples numerous solutions for each problem, then merges equivalent solution steps to construct a prefix tree (Trie), as shown in \Cref{fig:fig1}.
For each node in the tree, its Q-value is calculated as the proportion of solutions passing through it that reach the correct answer.
Each path from root to leaf in the tree represents a training sample with step-by-step labels.
Compared to simulation-based methods, SCOPE not only achieves $O(N)$ complexity through step compression, but also fully utilizes all sampled solutions by incorporating them directly into training data through the prefix tree structure.

The key challenge of SCOPE lies in identifying step equivalence. 
Naive exact string matching is too restrictive and results in limited compression.
Edit distance and sentence embeddings often fail to capture the subtle distinctions in mathematical reasoning \cite{wallace2019nlp}.
To address this challenge, we propose a code-based step compression through a three-stage process: (1) translate natural language reasoning steps into executable Python code using a code LLM, (2) normalize the code through Abstract Syntax Tree (AST) \cite{aho2007compilers} (e.g., variable renaming), and (3) merge steps with identical normalized code using a Trie. This approach enables precise identification of mathematically equivalent steps while being robust to surface-level variations. Although code translation and AST add computation, the overall complexity remains $O(N)$, ensuring substantially lower computational costs for large-scale PRM datasets.

Based on SCOPE, we construct a PRM training dataset containing 196K samples with 1.4M labels, exceeding Math-shepherd's scale while requiring only 5\% of its computational resources. Empirical evaluation demonstrates the effectiveness of our approach, with PRMs trained on our dataset consistently outperforming other automated annotation approaches in both the Best-of-N strategy and the ProcessBench \cite{zheng2024processbench} evaluation.

Our main contributions are:
\begin{itemize}
    \item
        We propose SCOPE, a novel automatic PRM label annotation method that introduces a sample-and-compress paradigm to replace traditional sample-and-simulation paradigm, achieving $O(N)$ complexity.
    \item
        We construct a new PRM training dataset containing 196K samples and 1.4M step-level labels, while requiring only 5\% of the computational resources used by MathShepherd.
    \item
        Extensive experiments show that PRMs trained on our dataset consistently outperform other automated annotation approaches across multiple evaluation settings, including Best-of-N strategy and ProcessBench.
\end{itemize}

\begin{figure*}[t]
\centering
\includegraphics[width=1\textwidth]{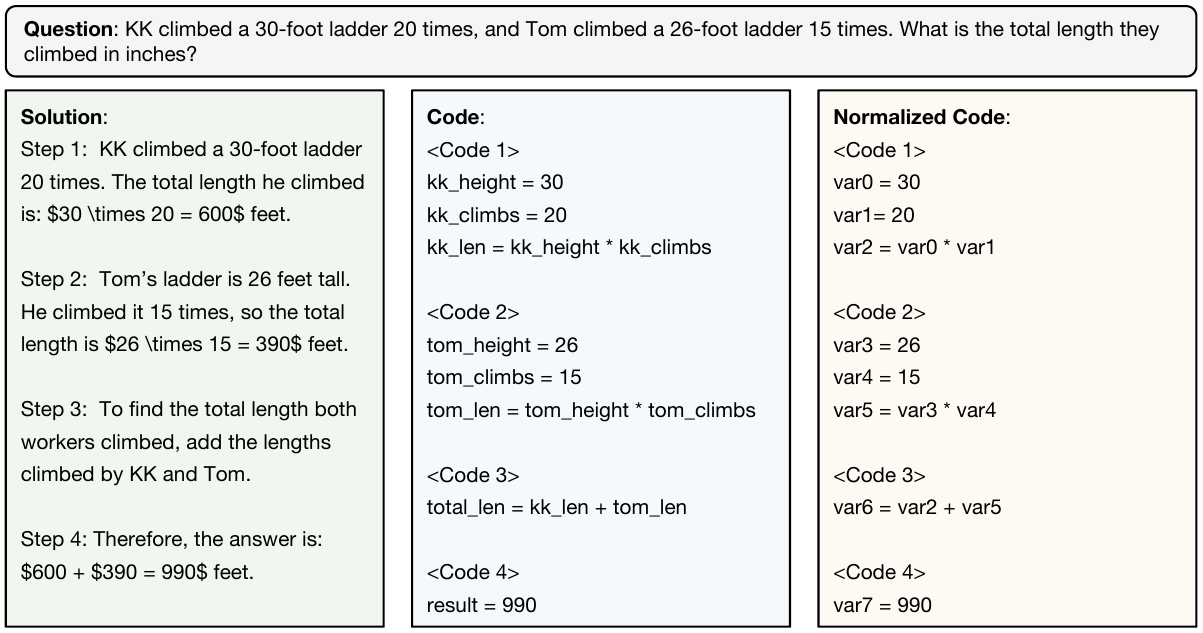}
\caption{
    Illustration of code translation and normalization.
    The solution of a math problem is first converted into corresponding codes through a code-LLM.
    Then, we use AST module of Python to derive the abstract syntax tree.
    Finally, the codes are normalized via their corresponding AST.
}
\label{fig:fig2}
\end{figure*}

\section{Related Work}
\subsection{PRMs Training}
Process Reward Models (PRMs) demonstrate significant potential in mathematical reasoning tasks, though their traditional training approaches require substantial human annotation effort \cite{lightman2023let}. While Math-shepherd \cite{wang2024math} introduces an innovative approach using Monte Carlo simulation to automate PRM training, its practical applications are constrained by intensive computational demands. OmegaPRM \cite{luo2024improve} attempts to address these limitations through a divide-and-conquer Monte Carlo Tree Search strategy, yet computational costs remain a significant barrier. Recent research explores alternative approaches to reduce these computational requirements: ImplicitPRM \cite{yuan2024free} demonstrates the possibility of deriving PRMs from outcome-level labels, while AutoPSV \cite{lu2024autopsv} develops a novel verification model that evaluates step quality through confidence variation analysis. However, recent studies \cite{zheng2024processbench} have revealed that these approaches often fail short of their claimed effectiveness, particularly struggling on more challenging datasets.

\subsection{PRMs in Mathematical Reasoning}
Process Reward Models enhance mathematical reasoning capabilities through dual mechanisms: reinforcement learning during the training phase and solution selection during inference. 
In reinforcement learning \cite{wang2024math,yuan2024free,shao2024deepseekmath}, PRMs serve as reward functions that guide policy optimization by providing fine-grained feedback on each reasoning step, enabling more targeted learning compared to traditional outcome-based rewards. 
During inference \cite{lu2024autopsv,lightman2023let,wang2024math}, the effectiveness of PRMs is commonly evaluated using the Best-of-N strategy, which identifies the highest-quality solution from multiple candidates by aggregating step-wise scores, demonstrating superior performance compared to outcome-based selection methods. 
The recent introduction of ProcessBench \cite{zheng2024processbench} establishes a more rigorous framework for evaluating PRMs' capabilities in identifying erroneous reasoning steps, offering a comprehensive assessment of their process-level understanding.
Building upon these insights into PRM effectiveness and evaluation frameworks, we evaluated SCOPE on both Best-of-N strategy and ProcessBench. 
Our method not only achieved state-of-the-art performance on Best-of-N, but also demonstrated remarkable effectiveness on the challenging ProcessBench.

\begin{figure*}[t]
\centering
\includegraphics[width=\textwidth]{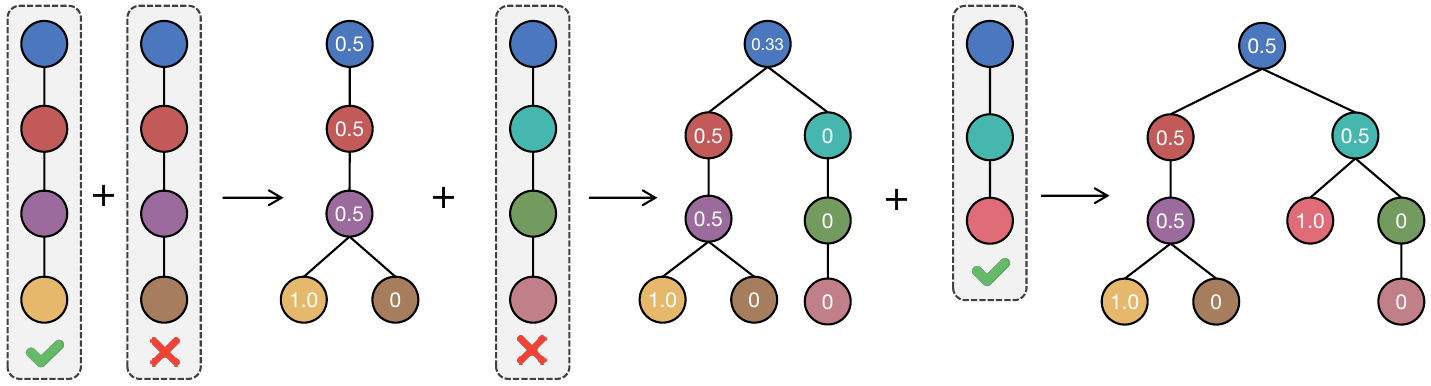}
\caption{Visualization of prefix tree construction process. Same-colored nodes indicate equivalent normalized step codes. Q-values reflect the proportion of correct solutions passing through each node.}
\label{fig:fig3}
\end{figure*}

\section{Method}
\label{sec:method}
In this section, we present SCOPE, a novel approach for automatic PRMs dataset annotation:
(1) First, we motivate and explain our code translation strategy, which converts reasoning steps into executable code through a code LLM. 
(2) Then, we perform AST normalization to standardize code syntax for accurate equivalence matching.
(3) Next, we apply step compression by merging normalized steps into a prefix tree.
(4) Finally, we describe our PRMs training details.

\subsection{Code Translation}
As discussed in Section \ref{sec:intro}, the key challenge of SCOPE lies in efficiently identifying and merging equivalent reasoning steps.
A naive approach based on exact string matching fails to recognize equivalent steps expressed differently (e.g., ``multiply 5 by 3'' vs ``calculate 5 × 3''), leading to an overly sparse compression space. 
While edit distance or sentence embeddings offer more flexibility, they struggle with precise numerical comparisons or operator precedence (e.g., failing to distinguish between ``(3 + 4) × 2'' and ``3 + 4 × 2''), making them unreliable in mathematical scenario \cite{wallace2019nlp}.
The core issue is that they operate on surface-level text similarities rather than identifying true mathematical equivalence, which can manifest in various forms such as different arithmetic representations or algebraic transformations.

Therefore, we propose using code as an intermediate representation that can precisely capture mathematical operations and logical reasoning. 
As shown in \Cref{fig:fig2}, we first use the math LLM to sample $N$ solutions for each math problem, then employ a code LLM to convert each natural language step into executable code blocks (see prompt details in \Cref{app:code_prompt}).
This code translation transforms natural language steps into a more structured and precise representation, laying the foundation for identifying mathematical equivalence. A detailed example of complex code generation is provided in \Cref{app:code_examples}.

\subsection{AST Normalization}
While code translation captures mathematical operations precisely, direct code matching remains ineffective due to syntactic variations. 
For example, ``x = 5 * 3'' and ``result = 3 * 5'' would be treated as different operations despite being mathematically equivalent, due to different variable names and operand orders.
To address this, we utilize Abstract Syntax Tree (AST), which represents code as a hierarchical structure of its syntactic elements, to normalize code through systematic transformations:
\begin{itemize} %
    \item 
        Variable renaming: Mapping arbitrary variable names and function names to canonical form (e.g., var0, func0).
    \item 
        Operation normalization: Standardizing equivalent operations (e.g., multiply/times/product $\rightarrow$ mul).
    \item 
        Expression reordering: Sorting commutative operations for consistent representation.
    \item 
        Constant folding: calculating constant expressions (e.g., 2 * 3 $\rightarrow$ 6).
\end{itemize}
Through AST normalization, as shown in Figure \ref{fig:fig2}, original code (middle column) becomes standardized (right column), enabling precise matching of equivalent steps. The complete AST structure for this code example is provided in Appendix \ref{app:ast}.

\subsection{Step Compression}
After AST normalization, we identify identical normalized code blocks and merge them to construct a prefix tree (Trie), where each node represents a distinct solution step and edges denote reasoning branches, as shown in Figure \ref{fig:fig3}. Note that some steps may contain only comments without code (see example in Appendix \ref{app:code_examples}). We retain such steps as distinct nodes, which account for 27.3\% of all steps. Section \ref{comp_strategy} discusses how different handling strategies affect the results.

This hierarchical representation enables efficient Q-value computation directly from the solution paths without requiring additional Monte Carlo simulations.
The Q-value is calculated recursively by propagating the correctness of leaf solutions up through the tree, weighted by the number of solutions passing through each path.
The pseudocode for Q-value calculation is shown below. 

\begin{lstlisting}[language=Python, basicstyle=\ttfamily\small]
def compute_q(node):
  if node.is_leaf():
    return node.is_correct, node.count
    
  total_value, total_count = 0, 0
  for child in node.children:
    value, count = compute_q(child)
    total_value += value * count
    total_count += count
    
  q_value = total_value / total_count
  return q_value, total_count
\end{lstlisting}

Since these three stages have linear complexity to the number of solutions, SCOPE maintains an overall complexity of $O(N)$. A detailed explanation is provided in Appendix \ref{app:complex_ana}.

\clearpage
\subsection{PRMs Training}
\label{sec:prm_training}
Through the above step compression process, we obtain Q-values for each step, which naturally serve as labels for their corresponding reasoning steps.
Following Math-shepherd \cite{wang2024math}, we explore two strategies to estimate the label $y_{s_i}$ for each step $s_i$, hard estimation (HE) and soft estimation (SE).
For HE, we assign binary labels based on the Q-value $Q(s_i)$ of step $s_i$: a positive Q-value indicates that at least one solution path through this step reaches the correct answer:
\begin{equation}
y_{s_i}^{HE} = \begin{cases}
1 & Q(s_i) > 0 \\
0 & \text{Otherwise}
\end{cases}
\end{equation}
For SE, we directly use the Q-value as the label, which reflects the proportion of paths from this step that reach the correct answer:
\begin{equation}
y_{s_i}^{SE} = Q(s_i)
\end{equation}

We adopt different loss functions for HE and SE to align with their respective label characteristics. For HE with binary labels, we use the binary cross-entropy loss for optimization:
\begin{equation}
\mathcal{L}_{\text{HE}} = -\sum_{i=1}^{K} y_{s_{i}} \log \hat{y}_{s_{i}} + (1 - y_{s_{i}}) \log (1 - \hat{y}_{s_{i}})
\end{equation}
where $\hat{y}_{s{i}}$ is the model's predicted probability for step $s_i$, and $K$ is the total number of steps. For SE with continuous Q-values as labels, we employ the mean squared error (MSE) loss:
\begin{equation}
\mathcal{L}_{\text{SE}} = -\sum_{i=1}^{K} (y_{s_i} - \hat{y}_{s_i})^2
\end{equation}
The choice of different loss functions reflects the distinct nature of HE and SE: binary cross-entropy is suited for classification tasks with hard labels, while MSE better handles regression with continuous values.

\begin{table*}[t!]
\centering
\resizebox{\textwidth}{!}{
\begin{tabular}{lccccccc}
    \toprule
    \textbf{Setting} & \textbf{GSM8K} & \textbf{MATH} & \makecell[c]{\textbf{Minerva}\\\textbf{Math}} & \makecell[c]{\textbf{GaoKao}\\\textbf{2023 En}} & \makecell[c]{\textbf{Olympiad}\\\textbf{Bench}} & \makecell[c]{\textbf{College}\\\textbf{Math}} & \textbf{Avg.} \\
    \midrule
    Greedy & 95.5 & 83.0 & 34.6 & 64.2 & 38.2 & 46.3 & 60.3  \\
    \hdashline
    Pass@8 (Upper Bound) & 97.8 & 91.8 & 46.7 & 79.7 & 59.4 & 52.5 & 71.3 \\
    Majority@8 & 96.5 & 86.9 & 40.1 & 70.4 & 46.2 & 47.8 & 64.7  \\
    \midrule
    Math-Shepherd-PRM-7B & 96.2 & 81.7 & 33.8 & 63.6 & 39.1 & 45.5 & 60.0 \\
    RLHFlow-PRM-Mistral-8B & 96.0 & 85.7 & 37.5 & 70.9 & 43.3 & 47.6 & 63.5 \\
    RLHFlow-PRM-Deepseek-8B & 96.7 & 85.6 & 38.2 & 69.9 & 44.3 & 47.4 & 63.7 \\
    EurusPRM-Stage1 & 95.1 & 83.4 & 37.9 & 66.1 & 40.2 & 39.2 & 60.0  \\
    EurusPRM-Stage2 & 95.8 & 83.1 & 37.7 & 65.8 & 38.5 & 41.0 & 60.3 \\
    Skywork-PRM-1.5B & 96.2 & 86.3 & 38.2 & 70.6 & 43.0 & 48.1 & 63.7 \\ 
    Skywork-PRM-7B & 96.4 & 86.1 & \textbf{39.1} & 70.9 & 42.8 & 47.9 & 63.9 \\
    *Qwen2.5-Math-7B-PRM800K & 96.5 & 86.5 & 38.1 & 70.5 & 43.5 & 48.2 & 63.0 \\
    \midrule
    \rowcolor{lightpurple}
    \textbf{SCOPE} & \textbf{96.7} & \textbf{87.7} & 38.2 & 71.9 & \textbf{46.8} & \textbf{48.3} & \textbf{64.9} \\
    \hspace{1em}- w/o AST normalization & 96.5 & 87.3 & 38.2 & \textbf{72.2} & 45.6 & 48.3 & 64.7 \\
    \hspace{1em}- w/o code translation & 96.6 & 87.0 & 36.4 & 70.6 & 45.5 & 48.0 & 64.0 \\
    \hspace{1em}- Step Replacement & 96.6 & 87.3 & 37.1 & 71.0 & 45.3 & 48.3 & 64.3 \\
    \hspace{1em}- Step Skipping & 96.7 & 87.7 & 37.1 & 71.7 & 45.8 & 48.3 & 64.6\\
    \bottomrule
\end{tabular}}
\caption{Performance comparison on the Best-of-8 strategy. * is trained on high-quality manually annotated process-level data (PRM800K); all others are trained using automatically constructed datasets.}
\label{tab:bon}
\end{table*}

\section{Experiments}
All experiments are conducted on a server equipped with 8 NVIDIA A100-80GB GPUs and 512GB of system RAM.
We utilize PyTorch \cite{paszke2019pytorch} as the implementation framework, SGLang \cite{zheng2024sglang} for sampling and DeepSpeed \cite{aminabadi2022deepspeed} for distributed training.

\subsection{Settings}
\noindent
\textbf{Base Models.} 
For our experiments, we employ Qwen2.5-Math-7B-Instruct\footnote{\url{https://huggingface.co/Qwen/Qwen2.5-Math-7B-Instruct}} \cite{yang2024qwen2} as the base model for PRMs training and dataset construction.
For code translation, we utilize Qwen2.5-Coder-32B-Instruct\footnote{\url{https://huggingface.co/Qwen/Qwen2.5-Coder-32B-Instruct}} \cite{hui2024qwen2}, which exhibits strong performance in converting natural language into executable code. \\

\noindent
\textbf{Dataset Construction.} 
We construct our PRMs training dataset using the SCOPE pipeline. Starting from math problems in the UltraInteract dataset\footnote{\url{https://huggingface.co/datasets/openbmb/UltraInteract_sft}}, we generate 64 solutions per problem using Qwen2.5-Math-7B-Instruct. Following prior work, we remove problems whose model confidence\footnote{Defined as the proportion of correct solutions among the 64 generated by the base model.} equals 0 or 1, as such extremes may introduce training bias. We also observe that the distribution of model confidence has a strong impact on PRMs performance; thus, we retain only problems with model confidence greater than 0.75 to ensure label reliability. 

For each retained problem, Qwen2.5-Coder-32B-Instruct translates reasoning steps into code, followed by AST-based normalization. We then construct a prefix tree from the 64 normalized solutions and compute Q-values for each node. The final dataset contains 196K samples and 1.4M step labels. \\

\noindent
\textbf{Evaluation.}
We evaluated our approach using two complementary metrics:
(1) Consistent with previous work \cite{lightman2023let, luo2024improve}, we employ the \textbf{Best-of-N} (BoN) sampling strategy for evaluation, which selects the highest-scored response from $N$ candidates according to the PRM. 
Using Qwen2.5-Math-7B-Instruct, we sample $N=8$ responses across multiple mathematical benchmarks: GSM8K \cite{cobbe2021training}, MATH \cite{hendrycks2021measuring}, MinervaMath \cite{lewkowycz2022solving}, GaoKao2023En \cite{liao2024mario}, OlympiadBench \cite{he2024olympiadbench}, and CollegeMath \cite{tang2024mathscale}.
Each candidate solution is scored using the product of step-wise scores from the PRM.
(2) \textbf{ProcessBench} \cite{zheng2024processbench}, which is specifically designed to assess error identification in mathematical reasoning, contains four sub-benchmarks: GSM8K, MATH, OlympiadBench, and Omni-MATH \cite{gao2024omni}. 
ProcessBench requires models to either identify the first erroneous step in incorrect solutions or verify the correctnesss in valid solutions. \\

\begin{table*}[t!]
\centering
\renewcommand{\arraystretch}{1.1}
\resizebox{\textwidth}{!}{
\begin{tabular}{lcccccccccccccc}
    \toprule
    \multirow{2}{*}{\textbf{Model}} & \multicolumn{3}{c}{\textbf{GSM8K}} & \multicolumn{3}{c}{\textbf{MATH}} & \multicolumn{3}{c}{\textbf{OlympiadBench}} & \multicolumn{3}{c}{\textbf{Omni-MATH}} & \multirow{2}{*}{\textbf{Avg. F1}} \\
    \cmidrule{2-13}
    & Error & Correct & \textbf{F1} & Error & Correct & \textbf{F1} & Error & Correct & \textbf{F1} & Error & Correct & \textbf{F1} & \\
    \midrule
    Math-Shepherd-PRM-7B & 32.4 & 91.7 & 47.9 & 18.0 & 82.0 & 29.5 & 15.0 & 71.1 & 24.8 & 14.2 & 73.0 & 23.8 & 31.5 \\
    RLHFlow-PRM-Mistral-8B & 33.8 & 99.0 & 50.4 & 21.7 & 72.2 & 33.4 & 8.2 & 43.1 & 13.8 & 9.6 & 45.2 & 15.8 & 28.4 \\
    RLHFlow-PRM-Deepseek-8B & 24.2 & 98.4 & 38.8 & 21.4 & 80.0 & 33.8 & 10.1 & 51.0 & 16.9 & 10.9 & 51.9 & 16.9 & 26.6 \\
    Skywork-PRM-1.5B & 50.2 & 71.5 & 59.0 & 37.9 & 65.2 & 48.0 & 15.4 & 26.0 & 19.3 & 13.6 & 32.8 & 19.2 & 36.4 \\
    Skywork-PRM-7B & 61.8 & 82.9 & 70.8 & 43.8 & 62.2 & 53.6 & 17.9 & 31.9 & 22.9 & 14.0 & 41.9 & 21.0 & 42.1 \\
    EurusPRM-Stage1 & 46.9 & 42.0 & 44.3 & 33.3 & 38.2 & 35.6 & 23.9 & 19.8 & 21.7 & 21.9 & 24.5 & 23.1 & 31.2 \\
    EurusPRM-Stage2 & 51.2 & 44.0 & 47.3 & 36.4 & 35.0 & 35.7 & 25.7 & 18.0 & 21.2 & 23.1 & 19.1 & 20.9 & 31.3 \\
    *Qwen2.5-Math-7B-PRM800K & 53.1 & 95.3 & 68.2 & 48.0 & 90.1 & \textbf{62.6} & 35.7 & 87.3 & \textbf{50.7} & 29.8 & 86.1 & \textbf{44.3} & \textbf{56.5} \\
    \midrule
    \rowcolor{lightpurple}
    \textbf{SCOPE} & 59.9 & 86.0 & \textbf{70.6} & 50.8 & 74.9 & 60.6 & 35.7 & 59.3 & 44.6 & 31.4 & 62.2 & 41.7 & 54.4 \\
    \hspace{1em}- w/o AST Normalization & 61.4 & 85.5 & 71.4 & 51.9 & 74.6 & 61.2 & 34.3 & 56.6 & 42.8 & 29.8 & 57.3 & 39.2 & 53.6 \\
    \hspace{1em}- w/o Code Translation & 56.5 & 83.9 & 67.6 & 47.0 & 75.6 & 57.9 & 32.7 & 54.9 & 41.0 & 26.5 & 56.8 & 36.1 & 50.6 \\
    \hspace{1em}- Step Replacement & 53.6 & 87.6 & 66.5 & 35.5 & 83.0 & 49.8 & 23.8 & 69.3 & 35.4 & 16.5 & 68.5 & 26.6 & 44.6 \\
    \hspace{1em}- Step Skipping & 56.5 & 90.2 & 69.5 & 38.0 & 84.5 & 52.5 & 24.1 & 74.9 & 36.4 & 15.9 & 71.0 & 26.0 & 44.6 \\
    \bottomrule
\end{tabular}}
\caption{Performance comparison on ProcessBench. * is trained on high-quality manually annotated process-level data (PRM800K); all others are trained using automatically constructed datasets.}
\label{tab:pb}
\end{table*}

\noindent
\textbf{Baselines.}
We compare against 7B-scale PRMs:
Math-Shepherd-PRM-7B \cite{wang2024math} estimates process labels through Monte Carlo simulation. 
RLHFlow-PRM-Mistral-8B and RLHFlow-PRM-DeepSeek-8B \cite{xiong2024rlhflowmath} adopt Math-Shepherd's methodology with different optimization objectives. 
EurusPRM-Stage1 and EurusPRM-Stage2 \cite{cui2025process} learn process rewards implicitly from ORM-based training.
Skywork-PRM-1.5B and Skywork-PRM-7B \cite{skyworkopeno12024} are two recently released Qwen2.5-Math-based PRMs by Skywork.
Finally, we include Qwen2.5-Math-7B-PRM800K \cite{zheng2024processbench} as a strong baseline. It is fine-tuned on the PRM800K dataset \cite{lightman2023let}, a high-quality, manually annotated corpus of 265K process-level samples.

For fairness, we do not compare against Qwen2.5-Math-PRM \cite{zheng2024processbench}, which uses a 72B critic model to supervise 1.5M process-level annotations. Their focus is on distilling critic capabilities from large models, while our work aims to reduce annotation cost. \\

\noindent
\textbf{Training Details.} %
For solution generation, we set both the sampling temperature and top-p to 0.8, with a maximum new token limit of 2048 to ensure comprehensive solution generation. 
In the code translation phase, we employ a temperature of 0 to ensure deterministic outputs, maintaining a maximum new token limit of 2048.
For PRM training, we use a batch size of 256, gradient clipping of 1.0, and the AdamW optimizer \cite{loshchilov2017decoupled} with a learning rate of 5e-7 and warm-up ratio of 0.05. 

\subsection{Main Results}

\noindent
\textbf{Best-of-N.} Table \ref{tab:bon} presents a comprehensive comparison of our proposed SCOPE with existing PRMs on the Best-of-8 strategy. SCOPE achieves an average accuracy of 64.9\%, outperforming Math-Shepherd-PRM-7B by 4.9\% while requiring only 5\% of its computational cost. 
Notably, SCOPE also surpasses Qwen2.5-Math-7B-PRM800K (63.0\%), which is trained on high-quality, manually annotated process-level data, demonstrating the effectiveness of our approach.

For context, we also report three reference metrics: greedy decoding (60.3\%), majority voting (64.7\%), and pass@8 (71.3\%, upper bound). SCOPE is the only method that outperforms Majority@8, a very strong baseline that reflects the collective correctness of multiple model outputs. \\

\noindent
\textbf{ProcessBench.}
As a complementary evaluation metric, ProcessBench assesses PRMs' ability to either identify the first erroneous step in incorrect solution or verify the correctness of correct solution. As shown in Table \ref{tab:pb}, most existing PRMs trained on automatically annotated data struggle on this benchmark, with average F1 scores typically below 45\%.
The strongest performance is achieved by Qwen2.5-Math-7B-PRM800K (56.5\% F1), which is trained on a high-quality manually labeled process-level dataset (PRM800K). Our method, SCOPE, achieves 54.4\%, closely approaching this supervised upper bound, while being fully automatic.

\subsection{Ablation on Key Components}
To evaluate the impact of core components in SCOPE, we conduct two ablation studies: 
\begin{itemize}
    \item \textbf{w/o Code Translation}: This variant skips the code translation step and merges reasoning steps directly based on natural language.
    \item \textbf{w/o AST Normalization}: This variant performs code translation but skips AST-based normalization, merging code blocks in their raw form.
\end{itemize}

We begin by analyzing compression efficiency. We define the compression rate as the ratio of compressed nodes in the Trie to the total number of reasoning steps before compression. A higher compression rate indicates less effective compression, i.e., more redundancy remains. As shown in Figure \ref{fig:compression_rate}, w/o code translation leads to a high compression rate of 92.0\%, suggesting that natural language steps are too diverse to merge effectively. Introducing code translation reduces the rate to 79.5\%, while further applying AST normalization brings it down to 65.5\%.

We then examine the impact on downstream performance. From Table \ref{tab:bon} and Table \ref{tab:pb}, we observe: (1) removing code translation leads to a 0.9\% drop in Best-of-8 accuracy and a 3.8\% drop in ProcessBench F1. (2) Removing AST normalization also causes slight performance drops, indicating its positive contribution. These results confirm that both code translation and AST normalization are essential for improving compression quality and PRM performance.

\subsection{Comparison of Step Compression Strategies}
\label{comp_strategy}
Beyond evaluating the core components of SCOPE, we further investigate whether alternative strategies can improve the compression process, especially for comment-only steps. These steps, which account for 27.3\% of all reasoning steps, are treated as distinct nodes in our default setting. But is this the best design choice? We design two additional strategies:
\begin{itemize}
    \item \textbf{Step Replacement}: Replace all comment-only steps with a generic placeholder string like $step_i = "only comment"$. This allows more aggressive merging.
    \item \textbf{Step Skipping}: Omit comment-only steps from the Trie entirely. During Q-value computation, these steps inherit the score of their preceding step to preserve label continuity.
\end{itemize}                                                

\begin{figure}[t]
\centering
\includegraphics[width=\linewidth]{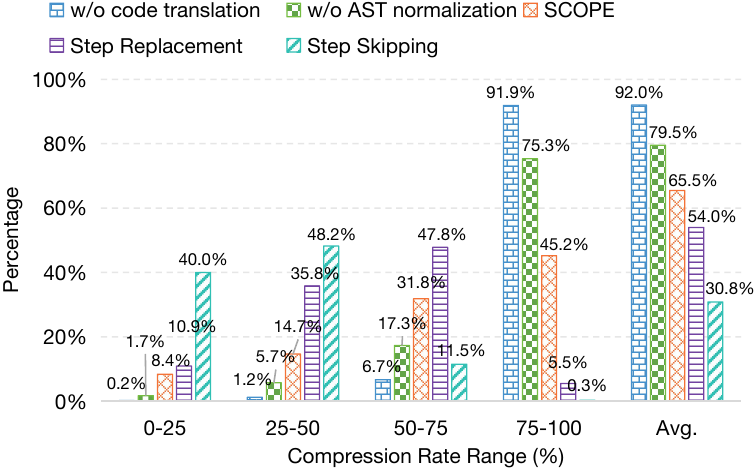}
\caption{Distribution of compression rates across different SCOPE variants. Compression Rate = compressed nodes in Trie / raw step count.}
\label{fig:compression_rate}
\end{figure}

Both strategies decrease the compression ratio. As shown in Figure \ref{fig:compression_rate}, Step Replacement and Step Skipping lead to compression rates of 54.0\% and 30.8\%, respectively—compared to 65.5\% in the default setting.
However, this decrease in compression ratio comes at a cost. As shown in Table \ref{tab:bon} and Table \ref{tab:pb}, both strategies result in lower performance on Best-of-8 and ProcessBench, despite achieving higher compression. These results suggest that over-compression harms PRM performance. Treating comment-only steps as distinct nodes, although computationally less efficient, provides better alignment supervision by preserving the reasoning structure.

\begin{table}[t]
\centering
\resizebox{0.9\linewidth}{!}{
\begin{tabular}{ccc}
\toprule
& Soft Label & Hard Label \\
\midrule
Best-of-8 (Avg. Acc) & 64.6 & 64.9 \\
ProcessBench (Avg. F1) & 52.8 & 54.4 \\
\bottomrule
\end{tabular}}
\caption{Comparison on soft and hard labels.}
\label{tab:label}
\end{table}

\subsection{Computational Efficiency}
\begin{figure}[t]
\centering
\includegraphics[width=0.9\linewidth]{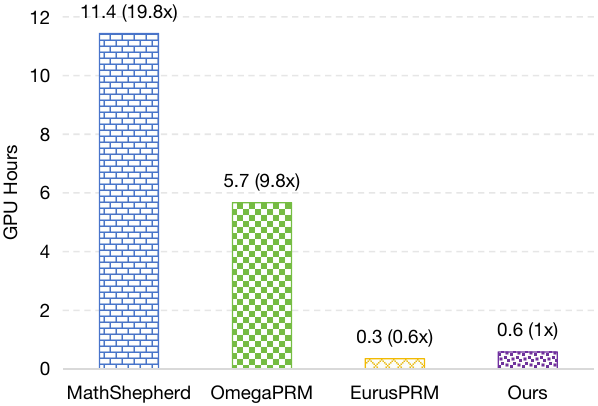}
\caption{Comparison of time costs (GPU hours) for generating PRM training data across different methods.}
\label{fig:flops}
\end{figure}

To evaluate the computational efficiency, we sample 100 problems from UltraInteract dataset and conduct PRMs training dataset using different methods. 
As shown in Figure \ref{fig:flops}, the GPU hours vary significantly across different approaches. MathShepherd \cite{wang2024math} requires 19.8$\times$ more GPU hours compared to our method. 
OmegaPRM \cite{luo2024improve} and EurusPRM \cite{cui2025process} consume 9.8$\times$ and 0.6$\times$ GPU hours respectively.
While EurusPRM shows faster computation, our previous experiments have demonstrated that it yields the poor performance on Best-of-8 and ProcessBench.
In contrast, our approach achieves strong performance while maintaining efficient computation. A detailed breakdown of computational costs for different stages in our method is provided in \Cref{app:complex_ana}.

\subsection{Soft Labels vs. Hard Labels}
\label{svh}

As outlined in \Cref{sec:prm_training}, PRMs can be trained using either hard labels or soft labels. \Cref{tab:label} presents a comparative analysis of these two training approaches on both the Best-of-8 and ProcessBench benchmarks. The results consistently demonstrate the superiority of hard labels over soft labels across both evaluation settings, with hard labels achieving performance gains of 0.3\% and 1.6\% on Best-of-8 and ProcessBench respectively. We attribute the limited performance of soft labels to the noise they introduce into the training process. This limitation is particularly evident in complex math problems where correct intermediate steps might receive low soft labels due to the difficulty of reaching the correct final answer, which can introduce confusion during the training process.

\section{Conclusion}
This paper introduces SCOPE, a novel automatic PRMs dataset annotation method that significantly reduces computational costs while maintaining label quality. By translating natural language reasoning steps into code and merging equivalent steps through AST normalization, our approach achieves $O(N)$ complexity compared to previous $O(NMK)$ methods. Using only 5\% of the computational resources, we construct a large-scale dataset containing 196K samples, and PRMs trained on our dataset consistently outperform existing approaches on both Best-of-N strategy and ProcessBench, demonstrating SCOPE's effectiveness as a scalable solution for PRM training.

\section*{Limitations}
While SCOPE demonstrates promising results, several limitations deserve attention:
(1) Code Translation Reliability: The quality of our annotations heavily depends on the code LLM's ability to accurately translate natural language reasoning into executable code. Complex mathematical concepts or domain-specific terminology may lead to inaccurate translations, affecting the overall annotation quality.
(2) Limited Mathematical Coverage: Our current implementation primarily handles basic arithmetic operations and common mathematical functions. More sophisticated mathematical operations, especially those involving abstract algebra or advanced calculus, may not be adequately captured by our code-based representation.

Future work could focus on developing more robust code translation techniques and expanding the coverage of mathematical operations. Additionally, investigating the integration of domain-specific knowledge \cite{pan2024fallacy,wu2020short,wu2024survey,wu2024akew,wu2024antileak} and mathematical formalism could further improve the accuracy and applicability of our approach.

\bibliography{custom}

\begin{thebibliography}{37}
\providecommand{\natexlab}[1]{#1}

\bibitem[{Aho et~al.(2007)Aho, Lam, Sethi, and Ullman}]{aho2007compilers}
Alfred Aho, Monica Lam, Ravi Sethi, and Jeffrey~D Ullman. 2007.
\newblock \href {https://dl.acm.org/doi/10.5555/1177220} {Compilers: Principles, techniques and tools, 2nd editio}.

\bibitem[{Aminabadi et~al.(2022)Aminabadi, Rajbhandari, Awan, Li, Li, Zheng, Ruwase, Smith, Zhang, Rasley et~al.}]{aminabadi2022deepspeed}
Reza~Yazdani Aminabadi, Samyam Rajbhandari, Ammar~Ahmad Awan, Cheng Li, Du~Li, Elton Zheng, Olatunji Ruwase, Shaden Smith, Minjia Zhang, Jeff Rasley, et~al. 2022.
\newblock \href {https://arxiv.org/abs/2207.00032} {Deepspeed-inference: enabling efficient inference of transformer models at unprecedented scale}.
\newblock In \emph{SC22: International Conference for High Performance Computing, Networking, Storage and Analysis}, pages 1--15. IEEE.

\bibitem[{Cobbe et~al.(2021)Cobbe, Kosaraju, Bavarian, Chen, Jun, Kaiser, Plappert, Tworek, Hilton, Nakano et~al.}]{cobbe2021training}
Karl Cobbe, Vineet Kosaraju, Mohammad Bavarian, Mark Chen, Heewoo Jun, Lukasz Kaiser, Matthias Plappert, Jerry Tworek, Jacob Hilton, Reiichiro Nakano, et~al. 2021.
\newblock \href {https://arxiv.org/abs/2110.14168} {Training verifiers to solve math word problems}.
\newblock \emph{arXiv preprint arXiv:2110.14168}.

\bibitem[{Cui et~al.(2025)Cui, Yuan, Wang, Wang, Li, He, Fan, Yu, Xu, Chen et~al.}]{cui2025process}
Ganqu Cui, Lifan Yuan, Zefan Wang, Hanbin Wang, Wendi Li, Bingxiang He, Yuchen Fan, Tianyu Yu, Qixin Xu, Weize Chen, et~al. 2025.
\newblock \href {https://arxiv.org/abs/2502.01456} {Process reinforcement through implicit rewards}.
\newblock \emph{arXiv preprint arXiv:2502.01456}.

\bibitem[{Dubey et~al.(2024)Dubey, Jauhri, Pandey, Kadian, Al-Dahle, Letman, Mathur, Schelten, Yang, Fan et~al.}]{dubey2024llama}
Abhimanyu Dubey, Abhinav Jauhri, Abhinav Pandey, Abhishek Kadian, Ahmad Al-Dahle, Aiesha Letman, Akhil Mathur, Alan Schelten, Amy Yang, Angela Fan, et~al. 2024.
\newblock \href {https://arxiv.org/abs/2407.21783} {The llama 3 herd of models}.
\newblock \emph{arXiv preprint arXiv:2407.21783}.

\bibitem[{Gao et~al.(2024)Gao, Song, Yang, Cai, Miao, Dong, Li, Ma, Chen, Xu et~al.}]{gao2024omni}
Bofei Gao, Feifan Song, Zhe Yang, Zefan Cai, Yibo Miao, Qingxiu Dong, Lei Li, Chenghao Ma, Liang Chen, Runxin Xu, et~al. 2024.
\newblock \href {https://arxiv.org/abs/2410.07985} {Omni-math: A universal olympiad level mathematic benchmark for large language models}.
\newblock \emph{arXiv preprint arXiv:2410.07985}.

\bibitem[{He et~al.(2024)He, Luo, Bai, Hu, Thai, Shen, Hu, Han, Huang, Zhang et~al.}]{he2024olympiadbench}
Chaoqun He, Renjie Luo, Yuzhuo Bai, Shengding Hu, Zhen~Leng Thai, Junhao Shen, Jinyi Hu, Xu~Han, Yujie Huang, Yuxiang Zhang, et~al. 2024.
\newblock \href {https://arxiv.org/abs/2402.14008} {Olympiadbench: A challenging benchmark for promoting agi with olympiad-level bilingual multimodal scientific problems}.
\newblock \emph{arXiv preprint arXiv:2402.14008}.

\bibitem[{Hendrycks et~al.(2021)Hendrycks, Burns, Basart, Zou, Mazeika, Song, and Steinhardt}]{hendrycks2021measuring}
Dan Hendrycks, Collin Burns, Saurav Basart, Andy Zou, Mantas Mazeika, Dawn Song, and Jacob Steinhardt. 2021.
\newblock \href {https://arxiv.org/abs/2103.03874} {Measuring mathematical problem solving with the math dataset}.
\newblock \emph{NeurIPS}.

\bibitem[{Hui et~al.(2024)Hui, Yang, Cui, Yang, Liu, Zhang, Liu, Zhang, Yu, Dang et~al.}]{hui2024qwen2}
Binyuan Hui, Jian Yang, Zeyu Cui, Jiaxi Yang, Dayiheng Liu, Lei Zhang, Tianyu Liu, Jiajun Zhang, Bowen Yu, Kai Dang, et~al. 2024.
\newblock \href {https://arxiv.org/abs/2409.12186} {Qwen2. 5-coder technical report}.
\newblock \emph{arXiv preprint arXiv:2409.12186}.

\bibitem[{Jaech et~al.(2024)Jaech, Kalai, Lerer, Richardson, El-Kishky, Low, Helyar, Madry, Beutel, Carney et~al.}]{jaech2024openai}
Aaron Jaech, Adam Kalai, Adam Lerer, Adam Richardson, Ahmed El-Kishky, Aiden Low, Alec Helyar, Aleksander Madry, Alex Beutel, Alex Carney, et~al. 2024.
\newblock \href {https://arxiv.org/abs/2412.16720} {Openai o1 system card}.
\newblock \emph{arXiv preprint arXiv:2412.16720}.

\bibitem[{Lewkowycz et~al.(2022)Lewkowycz, Andreassen, Dohan, Dyer, Michalewski, Ramasesh, Slone, Anil, Schlag, Gutman-Solo et~al.}]{lewkowycz2022solving}
Aitor Lewkowycz, Anders Andreassen, David Dohan, Ethan Dyer, Henryk Michalewski, Vinay Ramasesh, Ambrose Slone, Cem Anil, Imanol Schlag, Theo Gutman-Solo, et~al. 2022.
\newblock \href {https://arxiv.org/abs/2206.14858} {Solving quantitative reasoning problems with language models}.
\newblock \emph{Advances in Neural Information Processing Systems}, 35:3843--3857.

\bibitem[{Liao et~al.(2024)Liao, Luo, Li, Wu, and Fan}]{liao2024mario}
Minpeng Liao, Wei Luo, Chengxi Li, Jing Wu, and Kai Fan. 2024.
\newblock \href {https://arxiv.org/abs/2401.08190} {Mario: Math reasoning with code interpreter output--a reproducible pipeline}.
\newblock \emph{arXiv preprint arXiv:2401.08190}.

\bibitem[{Lightman et~al.(2023)Lightman, Kosaraju, Burda, Edwards, Baker, Lee, Leike, Schulman, Sutskever, and Cobbe}]{lightman2023let}
Hunter Lightman, Vineet Kosaraju, Yura Burda, Harri Edwards, Bowen Baker, Teddy Lee, Jan Leike, John Schulman, Ilya Sutskever, and Karl Cobbe. 2023.
\newblock \href {https://arxiv.org/abs/2305.20050} {Let's verify step by step}.
\newblock \emph{arXiv preprint arXiv:2305.20050}.

\bibitem[{Liu et~al.(2024)Liu, Feng, Xue, Wang, Wu, Lu, Zhao, Deng, Zhang, Ruan et~al.}]{liu2024deepseek}
Aixin Liu, Bei Feng, Bing Xue, Bingxuan Wang, Bochao Wu, Chengda Lu, Chenggang Zhao, Chengqi Deng, Chenyu Zhang, Chong Ruan, et~al. 2024.
\newblock \href {https://arxiv.org/abs/2412.19437} {Deepseek-v3 technical report}.
\newblock \emph{arXiv preprint arXiv:2412.19437}.

\bibitem[{Loshchilov(2017)}]{loshchilov2017decoupled}
I~Loshchilov. 2017.
\newblock \href {https://arxiv.org/abs/1711.05101} {Decoupled weight decay regularization}.
\newblock \emph{arXiv preprint arXiv:1711.05101}.

\bibitem[{Lu et~al.(2024)Lu, Dou, Hongru, Cao, Dai, Feng, and Guo}]{lu2024autopsv}
Jianqiao Lu, Zhiyang Dou, WANG Hongru, Zeyu Cao, Jianbo Dai, Yunlong Feng, and Zhijiang Guo. 2024.
\newblock \href {https://arxiv.org/abs/2405.16802} {Autopsv: Automated process-supervised verifier}.
\newblock In \emph{The Thirty-eighth Annual Conference on Neural Information Processing Systems}.

\bibitem[{Luo et~al.(2024)Luo, Liu, Liu, Phatale, Lara, Li, Shu, Zhu, Meng, Sun et~al.}]{luo2024improve}
Liangchen Luo, Yinxiao Liu, Rosanne Liu, Samrat Phatale, Harsh Lara, Yunxuan Li, Lei Shu, Yun Zhu, Lei Meng, Jiao Sun, et~al. 2024.
\newblock \href {https://arxiv.org/abs/2406.06592} {Improve mathematical reasoning in language models by automated process supervision}.
\newblock \emph{arXiv preprint arXiv:2406.06592}.

\bibitem[{o1~Team(2024)}]{skyworkopeno12024}
Skywork o1~Team. 2024.
\newblock \href {https://huggingface.co/Skywork} {Skywork-o1 open series}.
\newblock \url{https://huggingface.co/Skywork}.

\bibitem[{Pan et~al.(2024)Pan, Wu, Li, and Luu}]{pan2024fallacy}
Fengjun Pan, Xiaobao Wu, Zongrui Li, and Anh~Tuan Luu. 2024.
\newblock \href {https://doi.org/10.18653/v1/2024.emnlp-main.794} {Are {LLM}s good zero-shot fallacy classifiers?}
\newblock In \emph{Proceedings of the 2024 Conference on Empirical Methods in Natural Language Processing}, pages 14338--14364, Miami, Florida, USA. Association for Computational Linguistics.

\bibitem[{Paszke et~al.(2019)Paszke, Gross, Massa, Lerer, Bradbury, Chanan, Killeen, Lin, Gimelshein, Antiga et~al.}]{paszke2019pytorch}
Adam Paszke, Sam Gross, Francisco Massa, Adam Lerer, James Bradbury, Gregory Chanan, Trevor Killeen, Zeming Lin, Natalia Gimelshein, Luca Antiga, et~al. 2019.
\newblock \href {https://arxiv.org/abs/1912.01703} {Pytorch: An imperative style, high-performance deep learning library}.
\newblock \emph{Advances in neural information processing systems}, 32.

\bibitem[{Shao et~al.(2024)Shao, Wang, Zhu, Xu, Song, Bi, Zhang, Zhang, Li, Wu et~al.}]{shao2024deepseekmath}
Zhihong Shao, Peiyi Wang, Qihao Zhu, Runxin Xu, Junxiao Song, Xiao Bi, Haowei Zhang, Mingchuan Zhang, YK~Li, Y~Wu, et~al. 2024.
\newblock \href {https://arxiv.org/abs/2402.03300} {Deepseekmath: Pushing the limits of mathematical reasoning in open language models}.
\newblock \emph{arXiv preprint arXiv:2402.03300}.

\bibitem[{Snell et~al.(2024)Snell, Lee, Xu, and Kumar}]{snell2024scaling}
Charlie Snell, Jaehoon Lee, Kelvin Xu, and Aviral Kumar. 2024.
\newblock \href {https://arxiv.org/abs/2408.03314} {Scaling llm test-time compute optimally can be more effective than scaling model parameters}.
\newblock \emph{arXiv preprint arXiv:2408.03314}.

\bibitem[{Tang et~al.(2024)Tang, Zhang, Wang, and Wei}]{tang2024mathscale}
Zhengyang Tang, Xingxing Zhang, Benyou Wang, and Furu Wei. 2024.
\newblock \href {https://arxiv.org/abs/2403.02884} {Mathscale: Scaling instruction tuning for mathematical reasoning}.
\newblock \emph{arXiv preprint arXiv:2403.02884}.

\bibitem[{Uesato et~al.(2022)Uesato, Kushman, Kumar, Song, Siegel, Wang, Creswell, Irving, and Higgins}]{uesato2022solving}
Jonathan Uesato, Nate Kushman, Ramana Kumar, Francis Song, Noah Siegel, Lisa Wang, Antonia Creswell, Geoffrey Irving, and Irina Higgins. 2022.
\newblock \href {https://arxiv.org/abs/2211.14275} {Solving math word problems with process-and outcome-based feedback}.
\newblock \emph{arXiv preprint arXiv:2211.14275}.

\bibitem[{Wallace et~al.(2019)Wallace, Wang, Li, Singh, and Gardner}]{wallace2019nlp}
Eric Wallace, Yizhong Wang, Sujian Li, Sameer Singh, and Matt Gardner. 2019.
\newblock \href {https://arxiv.org/abs/1909.07940} {Do nlp models know numbers? probing numeracy in embeddings}.
\newblock \emph{arXiv preprint arXiv:1909.07940}.

\bibitem[{Wang et~al.(2024)Wang, Li, Shao, Xu, Dai, Li, Chen, Wu, and Sui}]{wang2024math}
Peiyi Wang, Lei Li, Zhihong Shao, Runxin Xu, Damai Dai, Yifei Li, Deli Chen, Yu~Wu, and Zhifang Sui. 2024.
\newblock \href {https://arxiv.org/abs/2312.08935} {Math-shepherd: Verify and reinforce llms step-by-step without human annotations}.
\newblock In \emph{Proceedings of the 62nd Annual Meeting of the Association for Computational Linguistics (Volume 1: Long Papers)}, pages 9426--9439.

\bibitem[{Wu(2025)}]{wu2025sailing}
Xiaobao Wu. 2025.
\newblock \href {https://arxiv.org/abs/2505.02686} {Sailing ai by the stars: A survey of learning from rewards in post-training and test-time scaling of large language models}.
\newblock \emph{arXiv preprint arXiv:2505.02686}.

\bibitem[{Wu et~al.(2020)Wu, Li, Zhu, and Miao}]{wu2020short}
Xiaobao Wu, Chunping Li, Yan Zhu, and Yishu Miao. 2020.
\newblock \href {https://aclanthology.org/2020.emnlp-main.138.pdf} {Short text topic modeling with topic distribution quantization and negative sampling decoder}.
\newblock In \emph{Proceedings of the 2020 Conference on Empirical Methods in Natural Language Processing (EMNLP)}, pages 1772--1782, Online.

\bibitem[{Wu et~al.(2024{\natexlab{a}})Wu, Nguyen, and Luu}]{wu2024survey}
Xiaobao Wu, Thong Nguyen, and Anh~Tuan Luu. 2024{\natexlab{a}}.
\newblock \href {https://doi.org/10.1007/s10462-023-10661-7} {A survey on neural topic models: Methods, applications, and challenges}.
\newblock \emph{Artificial Intelligence Review}.

\bibitem[{Wu et~al.(2024{\natexlab{b}})Wu, Pan, Wang, and Luu}]{wu2024akew}
Xiaobao Wu, Liangming Pan, William~Yang Wang, and Anh~Tuan Luu. 2024{\natexlab{b}}.
\newblock \href {https://doi.org/10.18653/v1/2024.emnlp-main.843} {{AKEW}: Assessing knowledge editing in the wild}.
\newblock In \emph{Proceedings of the 2024 Conference on Empirical Methods in Natural Language Processing}, pages 15118--15133, Miami, Florida, USA. Association for Computational Linguistics.

\bibitem[{Wu et~al.(2024{\natexlab{c}})Wu, Pan, Xie, Zhou, Zhao, Ma, Du, Mao, Luu, and Wang}]{wu2024antileak}
Xiaobao Wu, Liangming Pan, Yuxi Xie, Ruiwen Zhou, Shuai Zhao, Yubo Ma, Mingzhe Du, Rui Mao, Anh~Tuan Luu, and William~Yang Wang. 2024{\natexlab{c}}.
\newblock \href {https://arxiv.org/pdf/2412.13670} {{AntiLeak-Bench}: Preventing data contamination by automatically constructing benchmarks with updated real-world knowledge}.
\newblock \emph{arXiv preprint arXiv:2412.13670}.

\bibitem[{Xiong et~al.(2024)Xiong, Zhang, Jiang, and Zhang}]{xiong2024rlhflowmath}
Wei Xiong, Hanning Zhang, Nan Jiang, and Tong Zhang. 2024.
\newblock An implementation of generative prm.
\newblock \url{https://github.com/RLHFlow/RLHF-Reward-Modeling}.

\bibitem[{Yang et~al.(2024)Yang, Zhang, Hui, Gao, Yu, Li, Liu, Tu, Zhou, Lin, Lu, Xue, Lin, Liu, Ren, and Zhang}]{yang2024qwen2}
An~Yang, Beichen Zhang, Binyuan Hui, Bofei Gao, Bowen Yu, Chengpeng Li, Dayiheng Liu, Jianhong Tu, Jingren Zhou, Junyang Lin, Keming Lu, Mingfeng Xue, Runji Lin, Tianyu Liu, Xingzhang Ren, and Zhenru Zhang. 2024.
\newblock \href {https://arxiv.org/abs/2409.12122} {Qwen2.5-math technical report: Toward mathematical expert model via self-improvement}.
\newblock \emph{arXiv preprint arXiv:2409.12122}.

\bibitem[{Yu et~al.(2023)Yu, Gao, and Wang}]{yu2023outcome}
Fei Yu, Anningzhe Gao, and Benyou Wang. 2023.
\newblock \href {https://arxiv.org/abs/2311.09724} {Outcome-supervised verifiers for planning in mathematical reasoning}.
\newblock \emph{arXiv preprint arXiv:2311.09724}.

\bibitem[{Yuan et~al.(2024)Yuan, Li, Chen, Cui, Ding, Zhang, Zhou, Liu, and Peng}]{yuan2024free}
Lifan Yuan, Wendi Li, Huayu Chen, Ganqu Cui, Ning Ding, Kaiyan Zhang, Bowen Zhou, Zhiyuan Liu, and Hao Peng. 2024.
\newblock \href {https://arxiv.org/abs/2412.01981} {Free process rewards without process labels}.
\newblock \emph{arXiv preprint arXiv:2412.01981}.

\bibitem[{Zheng et~al.(2024{\natexlab{a}})Zheng, Zhang, Zhang, Lin, Lu, Yu, Liu, Zhou, and Lin}]{zheng2024processbench}
Chujie Zheng, Zhenru Zhang, Beichen Zhang, Runji Lin, Keming Lu, Bowen Yu, Dayiheng Liu, Jingren Zhou, and Junyang Lin. 2024{\natexlab{a}}.
\newblock \href {https://arxiv.org/abs/2412.06559} {Processbench: Identifying process errors in mathematical reasoning}.
\newblock \emph{arXiv preprint arXiv:2412.06559}.

\bibitem[{Zheng et~al.(2024{\natexlab{b}})Zheng, Yin, Xie, Sun, Huang, Yu, Cao, Kozyrakis, Stoica, Gonzalez et~al.}]{zheng2024sglang}
Lianmin Zheng, Liangsheng Yin, Zhiqiang Xie, Chuyue Sun, Jeff Huang, Cody~Hao Yu, Shiyi Cao, Christos Kozyrakis, Ion Stoica, Joseph~E Gonzalez, et~al. 2024{\natexlab{b}}.
\newblock \href {https://arxiv.org/abs/2312.07104} {Sglang: Efficient execution of structured language model programs}.
\newblock \emph{arXiv preprint arXiv:2312.07104}.

\end{thebibliography}

\clearpage

\appendix

\section{Code Translation Prompt}
\label{app:code_prompt}

\begin{minipage}{0.9\linewidth}
\setlength{\fboxsep}{10pt}
\colorbox{gray!10}{
  \begin{minipage}{\linewidth}
    \small
    You are a Python expert. I will provide a math problem along with a step-by-step solution. Please present each step of the solution as Python code. Ensure the following requirements are met: 

\begin{enumerate}[leftmargin=2em] %
    \item Clearly separate each step and save them in different code blocks, using<STEP\_START\_i> and <STEP\_END\_i> to separate them, where i represents the i-th step. 
    \item All calculations should be done in python code. Provide concise reasoning and thinking in the comments of the code. 
    \item If libraries are required, import them before the first step, using <IMPORT\_START> and <IMPORT\_END> tags. The most related python packages include `math', `sympy', `scipy', and `numpy'.
    \item Do not use any custom defined functions. Do implement the functionality with the simplest code.
    \item Ensure there is corresponding code for each step, even if the code is empty.
\end{enumerate}

\textbf{Math Problem:}\\
\texttt{...(math problem)...} \\

\textbf{Solution:}\\
\texttt{...(solution)...} \\
  \end{minipage}
}
\end{minipage}

\section{Complexity Analysis and Time Cost Evaluation}
\label{app:complex_ana}
In this appendix, we provide a more rigorous analysis of the theoretical time complexity of SCOPE, alongside a breakdown of the actual GPU an CPU time costs for each stage in our pipeline.

\subsection{Stage-wise Complexity Analysis}
SCOPE consists of the following stages: (1) Sampling of solutions, (2) Code translation, (3) AST normalization, and (4) Step compression. We analyze the complexity of each component below:

\begin{itemize}
    \item \textbf{Solution Sampling.} For each math problem, we sample $N$ complete solutions using a mathematical LLM. Since each solution is generated in a single forward pass, the overall sampling complexity is $O(N)$.
    \item \textbf{Code Translation.} Each solution is translated into a sequence of code steps using a code LLM in a single call. Although this is the most time-consuming stage in practice due to the use of a large LLM, the complexity remains $O(N)$, as each solution corresponds to one model invocation.
    \item \textbf{AST Normalization.} After code translation, each of the $K$ steps in all $N$ solutions is normalized using AST-based transformations. As each normalization call processes a complete solution at once, this stage also has linear complexity $O(N)$. Moreover, it runs entirely on CPU and typically completes in under 10 seconds, making its cost negligible in practice.
    \item \textbf{Step Compression.} Normalized code sequences are merged into a prefix tree (Trie). Trie construction requires a single pass through all steps, resulting time complexity of $O(N)$. In practice, this stage also runs on CPU and completes within 5 seconds.
\end{itemize}

All stages in SCOPE have linear time complexity with respect to the number of sampled solutions. Therefore, the end-to-end complexity of SCOPE remains $O(N)$.

\subsection{Actual Time Cost Breakdown}
Table \ref{tab:time_cost} reports time complexity and empirical GPU hours required for each stage of our method.

\begin{table*}[htbp!]
\centering
\begin{tabular}{lccc}
\toprule
Stage & Time Complexity & GPU Hours & Comment \\
\midrule
Solution Sampling & $O(N)$ & 192 & Using Qwen2.5-Math-7B \\
Code Translation & $O(N)$ & 280 & Using Qwen2.5-Coder-32B \\
AST Normalization & $O(N)$ & \textasciitilde0 & < 10 seconds on CPU \\
Step Compression & $O(N)$ & \textasciitilde0 & < 5 seconds on CPU \\
Total & $O(N)$ & \textasciitilde542 & Efficient end-to-end pipeline \\
\bottomrule
\end{tabular}
\caption{Breakdown of time complexity and costs in different stages of our method.}
\label{tab:time_cost}
\end{table*}

\section{AST Structure}
\label{app:ast}
\begin{figure}[htbp]
\centering
\includegraphics[width=\linewidth]{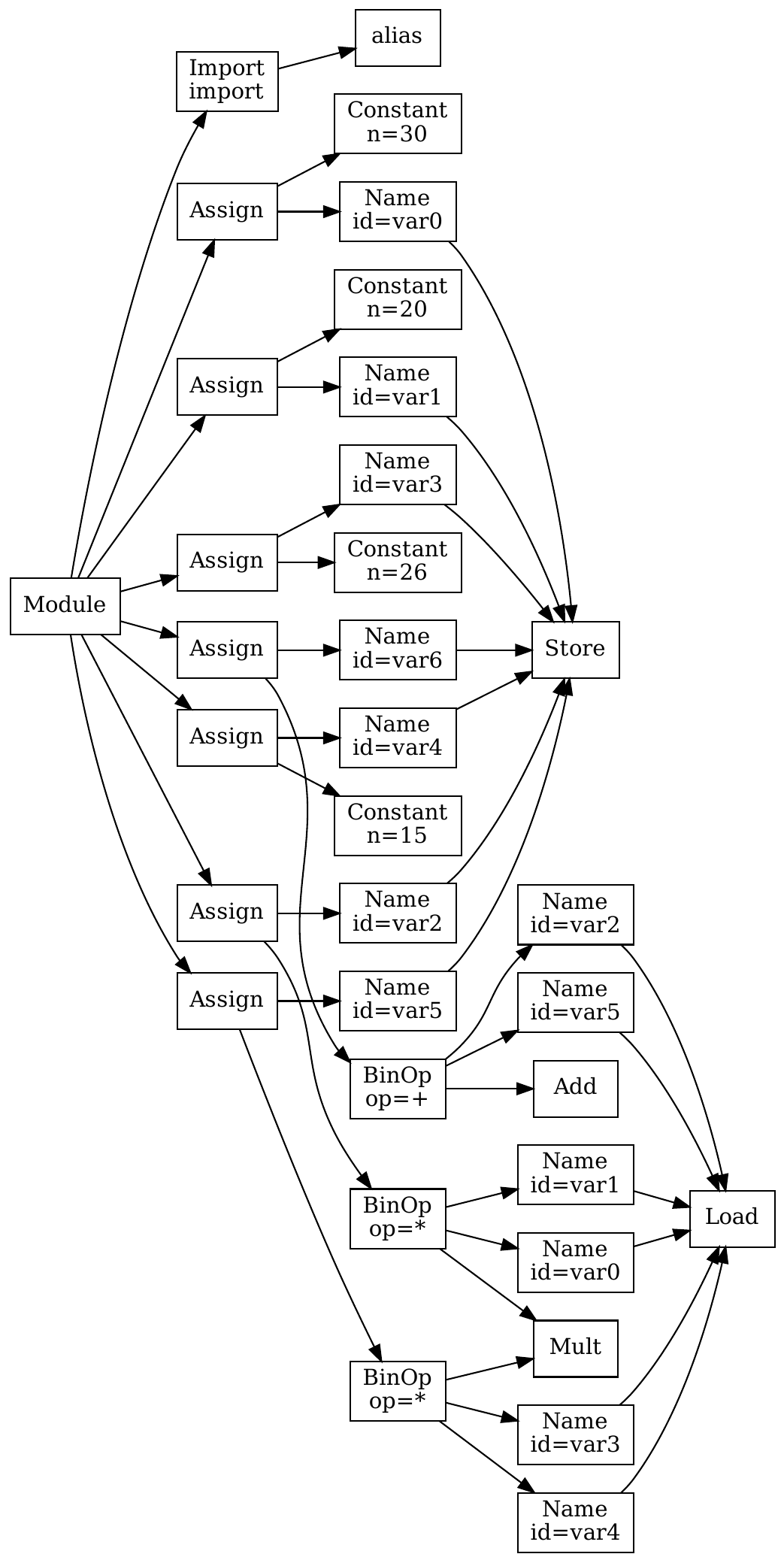}
\caption{AST structure of the ladder problem code.}
\label{fig:ast}
\end{figure}

Figure \ref{fig:ast} illustrates the Abstract Syntax Tree (AST) representation used in our code normalization process. The AST transforms code into a hierarchical structure that prioritizes semantic relationships over syntactic details.
In this representation, the root Module node connects to various Assign operations that define variables. Each variable assignment includes nodes for the target identifier (Name with id attribute) and its value, which may be a Constant or a binary operation (BinOp). Binary operations are represented with explicit operator nodes (Add, Mult) connecting their operands. 
This AST-based approach enables our system to recognize that statements like ``kk\_len = kk\_height * kk\_climbs'' and ``len = height * climbs'' are structurally equivalent after normalization, forming the foundation for our step compression algorithm.

\section{A Complex Code Translation Example}
\label{app:code_examples}
Table \ref{tab:example} shows a code translation example for a complex problem, where three steps are comments only.

\begin{table*}[htbp]
\resizebox{\textwidth}{!}{
\begin{tabular}{p{0.5\textwidth}|p{0.5\textwidth}}
\toprule
\multicolumn{2}{p{\textwidth}}{\textbf{Question:} The quadratic $8x^2+12x-14$ has two real roots. What is the sum of the squares of these roots? Express your answer as a common fraction in lowest terms.} \\ \hline
\textbf{Response} & \textbf{Code} \\ \hline
\parbox[t]{0.5\textwidth}{%
\textbf{Step 1}: We are given the quadratic equation $8x^2 + 12x - 14$ and are asked to find the sum of the squares of its roots. \\

\textbf{Step 2}: To find the sum of the squares of the roots, we can use the relationship between the sum of the roots, the product of the roots, and the coefficients of the quadratic equation. \\

\textbf{Step 3}: Let's denote the roots of the quadratic equation as $x_1$ and $x_2$. We know that the sum of the roots of a quadratic equation $ax^2+bx+c=0$ is given by $x_1 + x_2 = -\frac{b}{a} = -\frac{12}{8}$ and the product of the roots is given by $x_1x_2 = \frac{c}{a} = \frac{-14}{8} = -\frac{7}{4}$. \\

\textbf{Step 4}: In order to find the sum of the squares of the roots, which is given by $x_1^2 + x_2^2$, we can use the identity $(x_1 + x_2)^2 = x_1^2 + 2x_1x_2 + x_2^2$ to express the sum of the squares of the roots in terms of the sum and product of the roots. \\

\textbf{Step 5}: Substituting the values we found earlier, we get $(x_1 + x_2)^2 = \left(-\frac{3}{2}\right)^2 = \frac{9}{4}$. \\

\textbf{Step 6}: Then, we can express $x_1^2 + x_2^2$ as $(x_1 + x_2)^2 - 2x_1x_2$, which gives us $x_1^2 + x_2^2 = \frac{9}{4} - 2\left(-\frac{7}{4}\right)$. \\

\textbf{Step 7}: Simplifying this, we get $x_1^2 + x_2^2 = \frac{9}{4} + \frac{14}{4} = \frac{23}{4}$. \\

\textbf{Step 8}: Therefore, the sum of the squares of the roots of the quadratic equation $8x^2 + 12x - 14$ is $\frac{23}{4}$. }

&

\parbox[t]{0.5\textwidth}{%
\textbf{<CODE\_1>} \\
\# Given quadratic equation coefficients \\
a = 8 \\
b = 12 \\
c = -14 \\

\textbf{<CODE\_2>} \\
\# No calculations needed in this step, just setting up for next steps \\

\textbf{<CODE\_3>} \\
\# Calculate the sum and product of the roots \\
sum\_of\_roots = -b / a \\
product\_of\_roots = c / a \\

\textbf{<CODE\_4>} \\
\# No calculations needed in this step, just setting up for next steps \\

\textbf{<CODE\_5>} \\
\# Calculate the square of the sum of the roots \\
sum\_of\_roots\_squared = sum\_of\_roots ** 2 \\

\textbf{<CODE\_6>} \\
\# Calculate the sum of the squares of the roots using the identity \\
sum\_of\_squares\_of\_roots = sum\_of\_roots\_squared - 2 * product\_of\_roots \\

\textbf{<CODE\_7>} \\
\# Simplify the result \\
sum\_of\_squares\_of\_roots\_simplified = sp.Rational \\(sum\_of\_squares\_of\_roots).limit\_denominator() \\

\textbf{<CODE\_8>} \\
\# No calculations needed in this step, just stating the final answer }
\\
\bottomrule
\end{tabular}}
\caption{An example of natural language reasoning steps and their corresponding code translations for solving a quadratic equation problem.}
\label{tab:example}
\end{table*}

\end{document}